\documentclass[conference]{IEEEtran}
\IEEEoverridecommandlockouts
\usepackage{amsmath,amssymb,amsfonts}
\usepackage{algorithmic}
\usepackage{graphicx}
\usepackage{textcomp}
\usepackage{xcolor}
\usepackage{array}
\usepackage{mdwmath}
\usepackage{mdwtab}
\usepackage{eqparbox}
\usepackage{url}
\usepackage{hyperref}
\usepackage{tabularx}
\usepackage{subcaption}
\hypersetup{citecolor=Green}
\usepackage{mathrsfs}
\usepackage{rsfso}
\usepackage{multirow}
\usepackage{amssymb}
\usepackage{lipsum}
\usepackage{subfloat}
\usepackage[style=ieee, sorting=none]{biblatex}
\usepackage{comment}

\def\BibTeX{{\rm B\kern-.05em{\sc i\kern-.025em b}\kern-.08em
    T\kern-.1667em\lower.7ex\hbox{E}\kern-.125emX}}

\makeatletter
 \let\old@ps@headings\ps@headings
\let\old@ps@IEEEtitlepagestyle\ps@IEEEtitlepagestyle
 \def\confheader#1{%
 
 \def\ps@IEEEtitlepagestyle{%
 \old@ps@IEEEtitlepagestyle%
 \def\@oddhead{\strut\hfill#1\hfill\strut}%
 \def\@evenhead{\strut\hfill#1\hfill\strut}%
 }%
 \ps@headings%
 }
 \makeatother

\begin{document}

\title{Location Agnostic Adaptive Rain Precipitation Prediction using Deep Learning
}

\author{
\IEEEauthorblockN{Md Shazid Islam\IEEEauthorrefmark{1}, Md Saydur Rahman\IEEEauthorrefmark{1}, Md Saad Ul Haque\IEEEauthorrefmark{2},\\Farhana Akter Tumpa\IEEEauthorrefmark{3}, Md Sanzid Bin Hossain\IEEEauthorrefmark{4}, Abul Al Arabi\IEEEauthorrefmark{5}}

\IEEEauthorblockA{\IEEEauthorrefmark{1}University of California Riverside, USA, \IEEEauthorrefmark{2}University of Florida, USA \\ \IEEEauthorrefmark{3}Ahsanullah University of Science and Technology, Bangladesh\\\IEEEauthorrefmark{4}University of Central Florida, USA,
\IEEEauthorrefmark{5} CSE, Texas A\&M University, USA\\
Email: misla048@ucr.edu, mrahm054@ucr.edu,
haque.m@ufl.edu,\\Tumpafarhanaakter@gmail.com,
md543636@ucf.edu,
abulalarabi@tamu.edu
}}


\maketitle

\begin{abstract}
Rain precipitation prediction is a challenging task as it depends on weather and meteorological features which vary from location to location.  As a result, a prediction model that performs well at one location does not perform well at other locations due to the distribution shifts.  In addition, due to global warming, the weather patterns are changing very rapidly year by year which creates the possibility of ineffectiveness of those models even at the same location as time passes. In our work, we have proposed an adaptive deep learning-based framework in order to provide a solution to the aforementioned challenges. Our method can generalize the model for the prediction of precipitation for any location where the methods without adaptation fail. Our method has shown $43.51 \%$, $ 5.09 \%$, and $ 38.62 \%$ improvement after adaptation using a deep neural network for predicting the precipitation of Paris, Los Angeles, and Tokyo, respectively. 
\end{abstract}




\section{Introduction}
Weather forecasting \cite{i1} plays an essential role in numerous applications related to hydrology such as flood control \cite{i2}, reduction of impact from disaster \cite{i3} from climate change, and so on. Accurate prediction of rain precipitation is very important for planning and decision-making to minimize loss from any potential occurrence of a disaster. However, predicting accurate rain precipitation is an arduous task due to the complex nature of weather features and the sparsity of weather data across different locations \cite{i4}. \\

The accessibility of meteorological and climate data gathered from a variety of sensors \cite{i5} and weather stations \cite{i6} has significantly increased due to new developments in data collection technologies. The possibility to increase the precision of rain precipitation forecast over time is made possible by this enormous volume of data. However, the availability and distribution of meteorological data vary greatly among geologically unique locations, making it challenging to create prediction models that can successfully generalize across various kinds of regions. Moreover, the recent climate change pattern has had a great impact on precipitation as global warming and rainfall are deeply inter wined. Furthermore, global warming can cause shifts in precipitation patterns, both in terms of spatial distribution and seasonal timing. While some regions may experience increased rainfall, others may be subject to frequent droughts\cite{i14}. \\

Recently, deep learning-based approaches have been very useful tools in multi-modal learning \cite{araf,i7}, optimization \cite{easy,power1,power2}, domain generalization \cite{i9}. Hence being inspired by  \cite{i10,i11,i12}, a potential solution to the challenges of precipitation prediction that are posed by diverse weather patterns in various regions can be data-driven generalization also known as domain adaptation \cite{i13}. Through the use of domain adaptation techniques \cite{power3}, models trained on one source domain can be adapted to perform successfully in a target domain where a significant domain shift is evident. Domain adaptation increases prediction accuracy by transferring important insights to the target domain where the pre-trained model of the source fails.

The application of domain adaptation in the context of rain precipitation prediction is a relatively new research area. Most existing studies in rain precipitation prediction have focused on specific locations or regions, neglecting the cross-domain differences in weather patterns. This limitation calls for the development of generalized adaptive rain precipitation prediction models that can effectively adapt to varying weather patterns without being limited to specific locations.

In this paper, we propose a novel deep learning-based precipitation prediction approach that can adapt to changes in weather features such as temperature, wind flow, humidity, and so on. Our approach aims to overcome the limitations of traditional prediction models that solely rely on target domain data by leveraging the knowledge and insights gained from various source domains. By adapting the prediction models to the target domain, we aim to improve the accuracy of rain precipitation prediction for diverse locations with different weather patterns.

The contributions of this study include the development of a domain adaptation framework specifically tailored for rain precipitation prediction and the exploration of location-agnostic adaptive models that can adapt to varying weather patterns. We conduct extensive experiments and evaluations to demonstrate the effectiveness of our proposed approach and compare it with alternative methods.

The remainder of this paper is organized as follows. Section \ref{rw} provides a comprehensive review of the related literature, discussing previous studies on rain precipitation prediction and domain adaptation. Section \ref{method} presents the methodology and framework of our proposed location-agnostic adaptive rain precipitation prediction approach. In Section \ref{exp_result}, we present the dataset, experimental setup, and results of our experiments.  Finally, Section \ref{conc} concludes the paper, highlighting the key findings and suggesting future research directions.

\section{Related Works}
\label{rw}
Several studies have focused on the development of techniques for adaptive rain precipitation prediction. These works take into account the issues posed by changing climates and the necessity for accurate forecasting in a variety of applications. This section provides an overview of relevant research that has explored different approaches to tackle these challenges.

Numerous research has been conducted to examine the effectiveness of various AI algorithms for rainfall prediction in specific areas. Pham et al.\cite{r7} investigated ANFIS, ANN, and SVM models for forecasting daily precipitation in Vietnam's Hoa Binh region, and reported that SVM produced the best forecasts across the models tested. Diez-Sierra and Jesus \cite{r9} evaluated  K-means clustering, K-nearest neighbors (K-NN), SVM, random-forest (RF), and NN for long-term forecasts of rainfall on the Spanish island of Tenerife, concluding that neural network models outperformed SVM, indicating their potential for adaptive rain precipitation prediction. 
Tang et al. \cite{r12} recently suggested a data augmentation strategy for expanding precipitation series data that is based on the K-means clustering algorithm and the synthetic minority oversampling technique (SMOTE). They assessed the performance of shallow ML models (RF, extreme gradient boosting) and deep learning models (recurrent neural network, long short-term memory) before and after the expansion of precipitation data from the Danjiangkou River Basin in China. Their findings indicated that data augmentation has the potential to improve the accuracy of medium- and long-term precipitation prediction.

Recent studies have investigated the use of hybrid models to predict precipitation. For example, Xiang et al. \cite{r3} suggested a hybrid model for short and long-term precipitation prediction in Bangladesh's northern region using support vector regression (SVR) and artificial neural network (ANN). In their work, they used an ensemble empirical mode decomposition (EEMD) \cite{r10} for early data decomposition. Di Nunno et al. \cite{r11} proposed a model for precipitation forecasting that incorporates M5P and SVR. Their hybrid model exhibited superior performance with high $R^2$ values, suggesting its efficacy in predicting precipitation. Danandeh Mehr et al. \cite{r8} examined monthly precipitation prediction in a semiarid region by integrating SVR with the Fire-Fly algorithm. The hybrid model was compared with SVR and multigene genetic programming (MGGP) algorithms. The results showed that the hybrid SVR-FFA model outperformed the standalone SVR, while the MGGP approach exceeded it. Utilizing pre-processing techniques such as discrete-wavelet-transform and seasonal decomposition, Tran Anh et al. \cite{r4} implemented a hybrid model to predict monthly precipitation. They incorporated two feed-forward neural networks, namely artificial neural networks and seasonal artificial neural networks, in their models and produced the most satisfactory simulation of rainfall prediction using wavelet transform.

Additionally, Ghamariadyan and Imteaz \cite{r5} developed a medium-term (1, 3, 6, and 12 months lead time) prediction model using a hybrid wavelet artificial neural network (WANN). Their comparative analysis shows that the WANN outperformed  ANN, ARIMA, multiple linear regression (MLR), and the current prediction system for Queensland, Australia, providing more precise rainfall predictions.

These related works demonstrate the significance of adopting various technique-based approaches, including machine learning models and hybrid methods, for adaptive rain precipitation prediction. The studies highlight the potential of these approaches in improving accuracy, computational efficiency, and the ability to capture different components and patterns in rainfall time series. Building upon these previous findings, this paper proposes a novel Location Agnostic Adaptive Rain Precipitation Prediction approach, aiming to address the challenges associated with changing climate conditions and improve the accuracy of rainfall forecasts across different locations.

\begin{figure*}
\centering

    \subfloat[Deep neural network] {{\includegraphics[width=   \columnwidth]{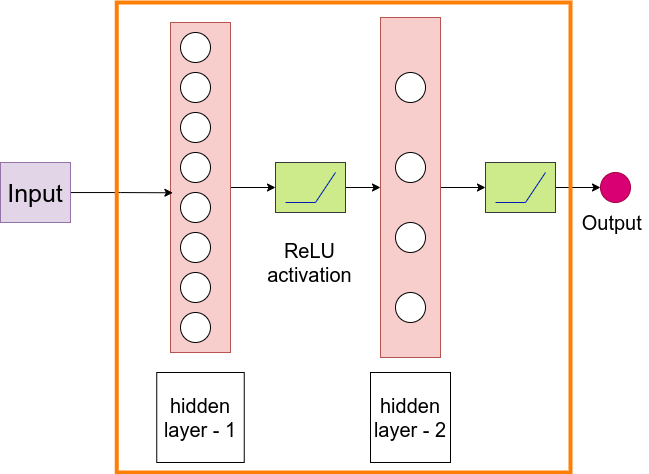}}
    \label{model}}
    \hfill
    \subfloat[Training on source side] {{\includegraphics[width=  \columnwidth]{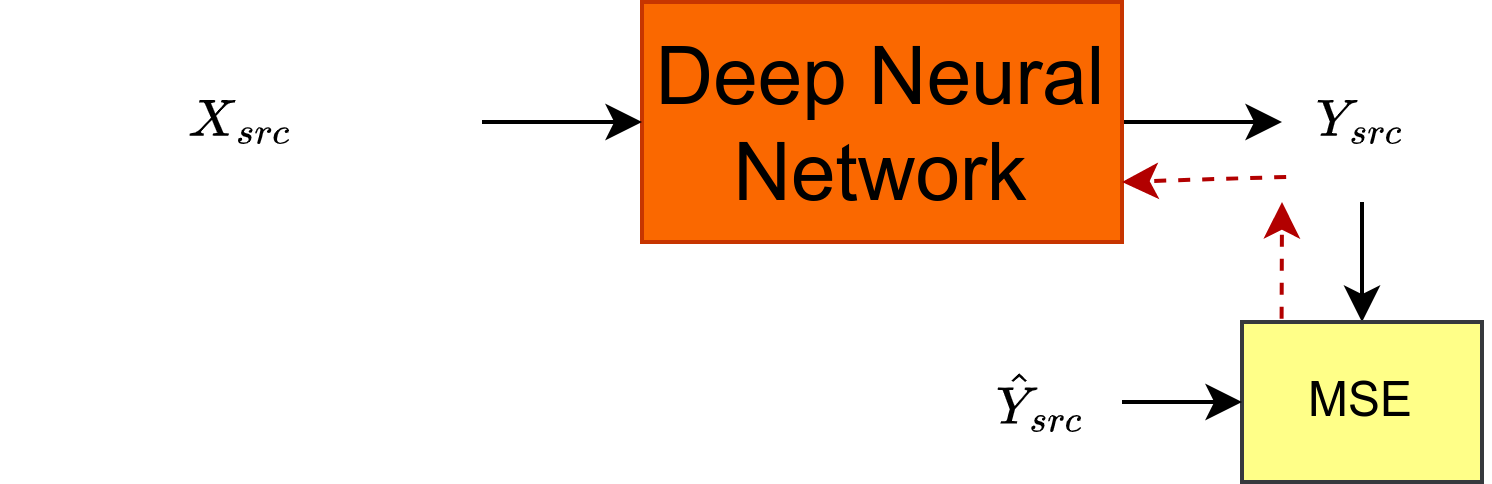}}
    \label{nw1}}

    \subfloat[Adaptation on target side]{{\includegraphics[width=  0.7\textwidth]{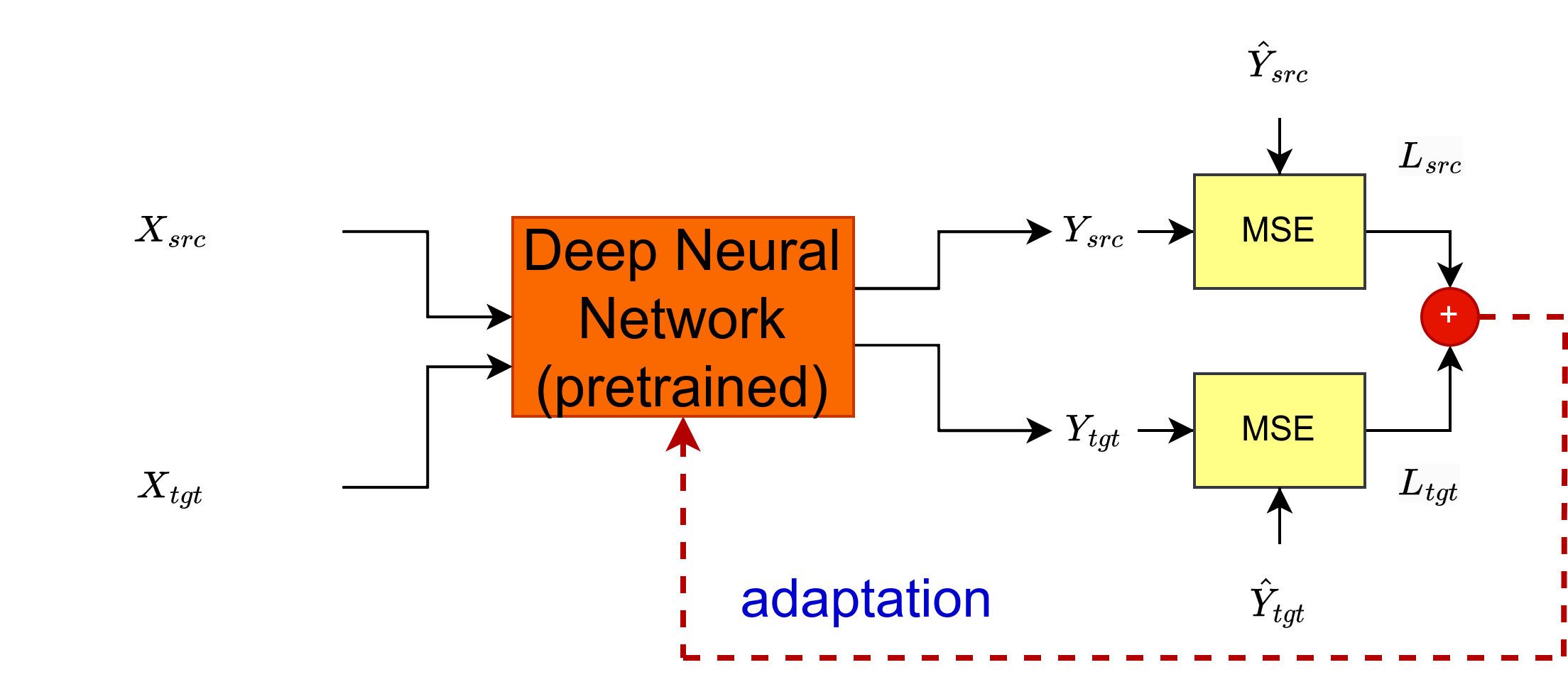}}
    \label{nw2}}

    \caption{The total workflow is shown in this figure. (a) The deep neural network that is used for training. We formulate our problem as a regression problem. Hence there is only one output node. (b) Training on the source side. $X_{src}$ is the source data (weather features) and $\hat{Y}_{src}$ is the ground truth data of the target value (precipitation). The training is guided by the mean square error (MSE) loss. (c) The adaptation on the target side is guided by MSE loss using both source and target data.}\label{workflow}
\end{figure*}

\section{Methodology}
\label{method}

In this paper, we use a deep learning model for rain precipitation prediction based on meteorological data as features. To make the model adaptive across spatial variation we leverage the domain adaptation approach. For domain adaptation, we need data from two domains which are termed as source domain and target domain. The source domain is the domain where the deep learning model will be trained primarily. The target domain is the domain where the trained model by the source data will be adapted. In our work, the source domain is the weather data of the city of Dhaka. The target domains are the weather data of cities -  Tokyo, Paris, and Los Angeles. In each experiment, only one of these cities will be considered as the target domain. The total workflow is shown in the Fig \ref{workflow}. At first, we shall train a deep learning network using the source data from scratch. Then we shall use data from both the source and target domain for adaptation.

\subsection{Training the Source Model}
We shall train a deep neural network using the source data. We know that neural networks can be trained for two types of problems which are classification and regression. We formulate our problem as a regression problem as we want to predict the value of precipitation.  A feed-forward deep neural network of Fig. \ref{model} comprising 2 stages of hidden layers is used for training. The output layer has one node in order to facilitate regression. If the output of the network (predicted precipitation) is $Y_{src}$ and the ground truth value is $\hat{Y}_{src}$, the mean square error (MSE) loss function can be expressed by 

\begin{equation}
L_{src} =    \frac{1}{N}\sum_{i=1}^{N} (Y_{src} - \hat{Y}_{src})^{2}
\end{equation}

where $N$ indicates the number of samples in training.

\subsection{Adaptation for Target Domain}

The target domain data is split into two portions, namely adaptation data and test data. The adaptation split of the target domain and the train split of the source domain are used for adapting the network for the target domain. The loss function which is used for the adaptation is expressed by

\begin{equation}
L_{total} =   \lambda_{1}L_{src} + \lambda_{2}L_{tgt}
\end{equation}

where $L_{src}$ and $L_{tgt}$ both are MSE loss and $\lambda_{1}$ and $\lambda_{2}$  are non negative weight coefficients.

\section{Experiments and Results}
\label{exp_result}

\subsection{Dataset}

Our findings from this dataset provide valuable insights into the relationships between different weather variables and offer valuable information for weather forecasting and understanding weather patterns in the Dhaka, Los Angeles, Paris, and Tokyo regions. We collected weather data from NASA Power \cite{b8} and conducted an extensive analysis of 20 years' worth of weather data, spanning from January 1, 2003, to January 1, 2023. The extracted features include air temperature, dew point, wet point, specific humidity, relative humidity, atmospheric pressure, and wind speed. (summarized in TABLE \ref{tab_feature} )

\begin{figure*}[t!]
\centering
\includegraphics[width=  0.6\textwidth]{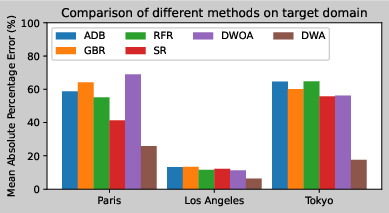}
\caption{Performance comparison on predicting precipitation of different methods by training them on the weather data of Dhaka city and using the pre-trained model for testing on Paris, Los Angeles, and Tokyo. ADB , GRB, RFR, SR, DWOA, and DWA indicate Adaboost, Gradient Boosting Regressor, Random Forest Regressor, Stacking Regressor, Deep learning WithOut Adaptation, and Deep learning With Adaptation, respectively. We see that Deep learning with adaptation outperforms all other methods by securing the lowest Mean Absolute percentage error.  }
\label{graphs}
\end{figure*}

\begin{table}[h!]
\caption{Features Used in the Dataset
}
\label{tab_feature}
\begin{center}
$\begin{array}{|c|c|}
\hline \textbf{ Features } & \begin{array}{c}
\textbf{ Description } 
\end{array} \\
\hline \begin{array}{c}
\text { T2M  } \\
\end{array} & \text{Average air temperature at 2 meters}  \\
\hline \begin{array}{c}
\text { T2MDEW} \\
\end{array} & \text{Dew point at 2 meters}\\
\hline \begin{array}{c}
\text {  T2MWET} \\
\end{array}& \text{Wet point at 2 meters}  \\
\hline \begin{array}{c}
\text { TS } \\
\end{array}& \text{Earth Skin Temperature} \\
\hline \begin{array}{c}
\text {T2M RANGE} \\

\end{array} & \text{ Merra-2 Temperature at 2 meters range} \\
\hline 
\begin{array}{c}
\text {  T2M MAX } \\
\end{array}& \text{Maximum temperature at 2 meters} \\
\hline 
\begin{array}{c}
\text {  T2M MIN } \\
\end{array}& \text{Minimum temperature at 2 meters} \\
\hline 
\begin{array}{c}
\text {  QV2M } \\
\end{array} &\text{Specific Humidity at 2 meters} \\
\hline 
\begin{array}{c}
\text {  RH2M } \\
\end{array}& \text{Relative Humidity at 2 meters} \\
\hline 
\begin{array}{c}
\text { PRECTOT } \\
\end{array} &\text{Bias-corrected total
precipitation at 2 meters} \\
\hline 
\begin{array}{c}
\text {  PS } \\
\end{array} &\text{Average surface pressure at surface} \\
\hline 
\begin{array}{c}
\text {  WS10M RANGE} \\
\end{array} &\text{Average wind speed 10 meters range} \\
\hline 
\begin{array}{c}
\text { WS10M } \\
\end{array} &\text{Average wind speed 10 meters high} \\
\hline 
\begin{array}{c}
\text {  WD10M } \\
\end{array} &\text{Average wind direction 10 meters high} \\
\hline 
\begin{array}{c}
\text {  WS10M MAX } \\
\end{array} &\text{maximum wind speed 10 meters} \\
\hline 
\begin{array}{c}
\text {  WS10M MIN } \\
\end{array} & \text{Minimum wind speed 10 meters}\\
\hline 

\end{array}$\\
\end{center}
\end{table}

\subsection{Results}
In the result section, we shall demonstrate the training performance of the deep neural network on source data, then show how the adaptive algorithm works in reducing mean square error and running time. \\

\subsubsection{Training Performance on Source Domain}
We present the performance of deep neural networks on training source data. We compare the performance of deep neural networks with some ensembling machine learning techniques such as Adaboost \cite{ab}, Gradient Boosting Regressor \cite{gbr}, Random Forest Regressor \cite{rfr} and  Stacking Regressor \cite{sr}. TABLE \ref{tab1} shows that Deep Neural Network performs the best (with the lowest MSE) compared to other ensembling machine learning techniques.
\\

\begin{table}
    \caption{Training on source data using different techniques}
    \begin{center}
        \begin{tabular}{|c|c|}
        \hline
        \textbf{Method}&{\textbf{Mean Square Error}} \\
        \hline
        Adaboost & 96.7630\\
        Gradient Boosting Regressor & 102.7857\\
        Random Forest Regressor & 85.8252\\
        Stacking Regressor & 112.3288\\
        Deep Neural Network & \textbf{79.8080}\\

        \hline

        \end{tabular}
    
    \label{tab1}
    \end{center}
\end{table}

\subsubsection{Performance of Adaptation on Target Domain} 
In our work, we have used `Dhaka' as our source domain. Tokyo, Los Angeles, and Paris have been selected as the target domain. The cities of the target domain have significant domain shifts compared to the source domain. TABLE \ref{tab2} shows when a deep neural network is trained on `Dhaka' and used in other cities without adaptation it exhibits high values of mean absolute error (MAE) in percentage. On the other hand, when adaptation is introduced, the MAE percentage decreases significantly. From TABLE \ref{tab2} we see, $43.51 \%$, $ 5.09 \%$, and $ 38.62  \%$ improvement in MAE for cities Paris, Los Angeles, and Tokyo, respectively. 
\begin{table}[h]
    \caption{Comparing the inference performance in terms of Mean Absolute Percentage Error for Deep Learning Network on other cities before and after adaptation. Here the city Dhaka is the source domain and Paris, Tokyo and Los Angeles are considered as the target domain. }
    \begin{center}
        \begin{tabular}{|c|c|c|}
        \hline
        \textbf{City} & Before Adaptation &After Adaptation   \\
        & MAE & MAE  \\
        
        \hline
        Paris & 69.0626 \%& \textbf{25.8855}\%\\

        Los Angeles & 11.3367 \% & \textbf{6.3358} \%\\
        
        Tokyo & 56.2065 \%& \textbf{17.5864} \%\\

        \hline

        \end{tabular}
    
    \label{tab2}
    \end{center}
\end{table}

Fig. \ref{graphs} illustrates how deep learning with adaptation on the target domain outperforms all other ensembling machine learning methods. It is evident that adaptation is essential if we want to generalize our model.

\section{Conclusion and Future Direction}
\label{conc}
This paper proposes a deep neural network-based adaptive technique in rain precipitation prediction. We have shown the comparison of different approaches on the target domain where our adaptive deep neural network outperforms other methods such as Adaboost, Gradient Boosting Regressor, Random Forest Regressor, and Stacking Regressor. Moreover the performance difference between before adaptation and after adaption is huge in the target domain area which satisfies our motivation. In the future, we will extend our work to fit the model using fewer features for different locations in order to make the approach computation and storage efficient. In addition, we shall focus on adaptation in an unsupervised manner.

\printbibliography

@article{i1,
  author = {Murphy, A.H.},
  title = {What is a good forecast? An essay on the nature of goodness in weather forecasting},
  journal = {Weather and Forecasting},
  volume = {8},
  number = {2},
  pages = {281-293},
  year = {1993}
}

@article{i2,
  author = {Pappenberger, F. and Thielen, J. and Del Medico, M.},
  title = {The impact of weather forecast improvements on large scale hydrology: analysing a decade of forecasts of the European Flood Alert System},
  journal = {Hydrological Processes},
  volume = {25},
  number = {7},
  pages = {1091-1113},
  year = {2011}
}

@article{i3,
  author = {Obasi, G.O.P.},
  title = {WMO's role in the international decade for natural disaster reduction},
  journal = {Bulletin of the American Meteorological Society},
  volume = {75},
  number = {9},
  pages = {1655-1661},
  year = {1994}
}

@article{i4,
  author = {Price, D.T. and McKenney, D.W. and Nalder, I.A. and Hutchinson, M.F. and Kesteven, J.L.},
  title = {A comparison of two statistical methods for spatial interpolation of Canadian monthly mean climate data},
  journal = {Agricultural and Forest Meteorology},
  volume = {101},
  number = {2-3},
  pages = {81-94},
  year = {2000}
}

@article{i5,
  author = {Rahut, Y. and Afreen, R. and Kamini, D. and Gnanamalar, S.S.},
  title = {Smart weather monitoring and real-time alert system using IoT},
  journal = {International Research Journal of Engineering and Technology},
  volume = {5},
  number = {10},
  pages = {848-854},
  year = {2018}
}

@article{i6,
  author = {Mendelsohn, R. and Kurukulasuriya, P. and Basist, A. and Kogan, F. and Williams, C.},
  title = {Climate analysis with satellite versus weather station data},
  journal = {Climatic Change},
  volume = {81},
  number = {1},
  pages = {71-83},
  year = {2007}
}

@inproceedings{i7,
  author = {Ngiam, Jiquan and others},
  title = {Multimodal deep learning},
  booktitle = {Proceedings of the 28th international conference on machine learning (ICML-11)},
  year = {2011}
}

@inproceedings{araf,
  author = {Islam, Md Shazid and Rahman, Md Saydur and Amin, M. Ashraful},
  title = {Beat Based Realistic Dance Video Generation using Deep Learning},
  booktitle = {2019 IEEE International Conference on Robotics, Automation, Artificial-intelligence and Internet-of-Things (RAAICON)},
  year = {2019}
}

@techreport{easy,
  title={Utilizing Genetic Evolution to Enhance Cellular Automata for Accurate Image Edge Detection},
  author={Tumpa, Farhana Akter and Rahman, Md Saydur and Islam, Md Shazid},
  year={2023},
  institution={EasyChair}
}

@inproceedings{power1,
  author={Jahid Hasan, A S M and Yusuf, Jubair and Rahman, Md Saydur and Islam, Md Shazid},
  booktitle={2023 International Conference on Information and Communication Technology for Sustainable Development (ICICT4SD)}, 
  title={Electricity Cost Optimization for Large Loads through Energy Storage and Renewable Energy}, 
  year={2023},
  volume={},
  number={},
  pages={46-50},
  doi={10.1109/ICICT4SD59951.2023.10303409}}

@inproceedings{power2,
  author={M Jahid Hasan, A S and Rahman, Md Saydur and Islam, Md Shazid and Yusuf, Jubair},
  booktitle={2023 International Conference on Information and Communication Technology for Sustainable Development (ICICT4SD)}, 
  title={Data Driven Energy Theft Localization in a Distribution Network}, 
  year={2023},
  volume={},
  number={},
  pages={388-392},
  doi={10.1109/ICICT4SD59951.2023.10303520}}

@inproceedings{power3,
  author = {Islam, Md Shazid and Hasan, A S M Jahid and Rahman, Md Saydur and Yousuf, Jubair and Sajol , Md Saiful Islam, and Tumpa, Farhana Akter},
  title = {Location Agnostic Source-Free Domain Adaptive Learning to Predict Solar Power Generation},
  booktitle = {2023 IEEE International Conference on Energy Technologies for Future Grids, Wollongong, Australia},
  year = {2023},
pages = {in press}
}

@article{i9,
  author = {Zhou, Kaiyang and others},
  title = {Domain generalization: A survey},
  journal = {IEEE Transactions on Pattern Analysis and Machine Intelligence},
  year = {2022}
}

@article{i10,
  author = {Wang, Mei and Deng, Weihong},
  title = {Deep visual domain adaptation: A survey},
  journal = {Neurocomputing},
  volume = {312},
  pages = {135-153},
  year = {2018}
}

@inproceedings{i11,
  author = {Farahani, Abolfazl and others},
  title = {A brief review of domain adaptation},
  booktitle = {Advances in data science and information engineering: proceedings from ICDATA 2020 and IKE 2020},
  year = {2021}
}

@article{i12,
  author = {Zhang, Siyu and others},
  title = {Rotating machinery fault detection and diagnosis based on deep domain adaptation: A survey},
  journal = {Chinese Journal of Aeronautics},
  volume = {36},
  number = {1},
  pages = {45-74},
  year = {2023}
}

@inproceedings{i13,
 
title={Generalized source-free domain adaptation},
  author={Yang, Shiqi and Wang, Yaxing and Van De Weijer, Joost and Herranz, Luis and Jui, Shangling},
  booktitle={Proceedings of the IEEE/CVF International Conference on Computer Vision},
  pages={8978--8987},
  year={2021}
}

@article{i14,
  author = {Lambert, F.H. and Stine, A.R. and Krakauer, N.Y. and Chiang, J.C.},
  title = {How much will precipitation increase with global warming?},
  journal = {EOS, Transactions American Geophysical Union},
  volume = {89},
  number = {21},
  pages = {193-194},
  year = {2008}
}

@article{r3,
  author = {Xiang, Y. and Gou, L. and He, L. and Xia, S. and Wang, W.},
  title = {A SVR–ANN combined model based on ensemble EMD for rainfall prediction},
  journal = {Applied Soft Computing},
  volume = {73},
  pages = {874-883},
  year = {2018}
}

@article{r4,
  author = {Tran Anh, D. and Duc Dang, T. and Pham Van, S.},
  title = {Improved rainfall prediction using combined pre-processing methods and feed-forward neural networks},
  journal = {J},
  volume = {2},
  number = {1},
  pages = {65-83},
  year = {2019}
}

@article{r5,
  author = {Ghamariadyan, M. and Imteaz, M.A.},
  title = {A Wavelet Artificial Neural Network method for medium‐term rainfall prediction in Queensland (Australia) and the comparisons with conventional methods},
  journal = {International Journal of Climatology},
  volume = {41},
  pages = {E1396-E1416},
  year = {2021}
}

@article{r7,
  author = {Pham, B.T. and Le, L.M. and Le, T.T. and Bui, K.T.T. and Le, V.M. and Ly, H.B. and Prakash, I.},
  title = {Development of advanced artificial intelligence models for daily rainfall prediction},
  journal = {Atmospheric Research},
  volume = {237},
  pages = {104845},
  year = {2020}
}

@article{r8,
  author = {Danandeh Mehr, A. and Nourani, V. and Karimi Khosrowshahi, V. and Ghorbani, M.A.},
  title = {A hybrid support vector regression–firefly model for monthly rainfall forecasting},
  journal = {International Journal of Environmental Science and Technology},
  volume = {16},
  pages = {335-346},
  year = {2019}
}

@article{r9,
  author = {Diez-Sierra, J. and Del Jesus, M.},
  title = {Long-term rainfall prediction using atmospheric synoptic patterns in semi-arid climates with statistical and machine learning methods},
  journal = {Journal of Hydrology},
  volume = {586},
  pages = {124789},
  year = {2020}
}

@article{r10,
  author = {Wang, W.c. and Chau, K.w. and Xu, D.m. and Chen, X.Y.},
  title = {Improving forecasting accuracy of annual runoff time series using ARIMA based on EEMD decomposition},
  journal = {Water Resources Management},
  volume = {29},
  pages = {2655–2675},
  year = {2015}
}

@article{r11,
  author = {Di Nunno, F. and Granata, F. and Pham, Q. and Marinis, G.},
  title = {Precipitation Forecasting in Northern Bangladesh Using a Hybrid Machine Learning Model},
  journal = {Sustainability},
  volume = {14},
  number = {5},
  year = {2022}
}

@article{r12,
  author = {Tang, T. and Jiao, D. and Chen, T. and Gui, G.},
  title = {Medium- and Long-Term Precipitation Forecasting Method Based on Data Augmentation and Machine Learning Algorithms},
  journal = {IEEE Journal of Selected Topics in Applied Earth Observations and Remote Sensing},
  volume = {15},
  pages = {1000-1011},
  year = {2022}
}

@online{b8,
  author = {P. Stackhouse},
  title = {NASA POWER},
  url = {https://power.larc.nasa.gov/dataaccess-viewer/},
  note = {Accessed: 2021-10-16}
}

@inproceedings{ab,
  author = {Schapire, Robert E.},
  title = {Explaining adaboost},
  booktitle = {Empirical Inference: Festschrift in Honor of Vladimir N. Vapnik},
  year = {2013}
}

@inproceedings{gbr,
  author = {Prettenhofer, Peter and Louppe, Gilles},
  title = {Gradient boosted regression trees in scikit-learn},
  booktitle = {PyData 2014},
  year = {2014}
}

@inproceedings{rfr,
  author = {Cootes, Tim F. and others},
  title = {Robust and accurate shape model fitting using random forest regression voting},
  booktitle = {Computer Vision–ECCV 2012: 12th European Conference on Computer Vision, Florence, Italy, October 7-13, 2012, Proceedings, Part VII 12},
  year = {2012}
}

@article{sr,
  author = {Acharya, Soumyadipta and others},
  title = {Non-invasive estimation of hemoglobin using a multi-model stacking regressor},
  journal = {IEEE Journal of Biomedical and Health Informatics},
  volume = {24},
  number = {6},
  pages = {1717-1726},
  year = {2019}
}

\vspace{12pt}
\color{red}

\end{document}